\pdfoutput=1

\documentclass[11pt]{article}

\usepackage[final]{coling}

\usepackage{times}
\usepackage{latexsym}
\usepackage{multirow}
\usepackage{booktabs}
\usepackage{amsmath}
\usepackage{xspace}
\usepackage[T1]{fontenc}

\usepackage[utf8]{inputenc}

\usepackage{microtype}

\usepackage{inconsolata}

\usepackage{graphicx}
\usepackage{hyperref}
\usepackage{makecell}
\usepackage{todonotes}
\usepackage[capitalise]{cleveref}
\usepackage{adjustbox}
\usepackage{comment}
\usepackage{tcolorbox}
\newcommand\F{{$\text{F}_1$}\xspace}

\title{HiTZ at VarDial 2025 NorSID: Overcoming Data Scarcity with Language Transfer and Automatic Data Annotation}

\author{Jaione Bengoetxea \quad Mikel Zubillaga \quad Ekhi Azurmendi \\ \bf Maite Heredia \quad Julen Etxaniz \quad Markel Ferro \quad Jeremy Barnes \\
HiTZ Center -- Ixa, University of the Basque Country (UPV/EHU) \\
\textit{name.surname}@ehu.eus \\
}

\begin{document}
\maketitle
\begin{abstract}
In this paper we present our submission for the NorSID Shared Task as part of the 2025 VarDial Workshop \cite{scherrer-etal-2025-vardial}, consisting of three tasks: Intent Detection, Slot Filling and Dialect Identification, evaluated using data in different dialects of the Norwegian language. 
For Intent Detection and Slot Filling, we have fine-tuned a multitask model in a cross-lingual setting, to leverage the xSID dataset available in 17 languages.
In the case of Dialect Identification, our final submission consists of a model fine-tuned on the provided development set, which has obtained the highest scores within our experiments. 
Our final results on the test set show that our models do not drop in performance compared to the development set, likely due to the domain-specificity of the dataset and the similar distribution of both subsets.
Finally, we also report an in-depth analysis of the provided datasets and their artifacts, as well as other sets of experiments that have been carried out but did not yield the best results. 
Additionally, we present an analysis on the reasons why some methods have been more successful than others; mainly the impact of the combination of languages and domain-specificity of the training data on the results.

\end{abstract}

\section{Introduction}
Dialectal variation is ubiquitous in human language and should be taken into account when performing Natural Language Processing (NLP) tasks, as NLP systems unable to deal with dialectal data can cause users to feel frustrated and lead to unintended biases \citep{Harwell-2018}. 

This is especially relevant for Spoken Language Understanding (SLU), a field of Speech Processing and Natural Language Understanding aimed at ensuring the semantic comprehension of human utterances by virtual assistants. To make systems that rely on SLU more robust and able to handle real use-cases, it is necessary to develop resources for these tasks not only for different languages, but for different language varieties, so that the benefits of these models can reach a wider variety of speech communities. 

With this motivation, the NorSID Shared Task consists of three subtasks (intent detection, slot filling and dialect identification) in four Norwegian variants: Bokmål (B), Western (V), Trøndersk (T) and North Norwegian (N). The tasks are centered around common virtual assistant tasks, such as setting alarms or questions about the weather.

Our team participated in all three subtasks, for a total of 6 runs: 3 for the SID (Slot and Intent Detection) tasks and 3 for Dialect Identification. As a team, we placed first in Dialect Identification, second in Intent Detection, and third in Slot Filling. Our code is publicly available on GitHub.\footnote{\url{https://github.com/hitz-zentroa/vardial-2025}}

\section{Task Descriptions}
As mentioned, this shared task consists of the following three subtasks:

\paragraph{Intent Detection.} It is a text classification task that assigns intent labels to the utterances of the users, to guide the chatbot's answer, depending on its domain and purpose.
\paragraph{Slot Filling.} It requires classifying token spans that contain relevant information for a virtual assistant to fulfill certain tasks, e.g., to set an alarm, the assistant needs to know the time to set it to. 
\paragraph{Dialect Identification.} The aim of this classification task is to identify the dialect of the utterance. 

{
\begin{table}[ht]
    \centering
    \small
    \begin{adjustbox}{max width=\linewidth}
    \begin{tabular}{lclll}
    \toprule
  id&text& intent&dialect &slots\\\midrule[1pt]
         90/9&Sett alarm for \colorbox{RosyBrown1}{kl. 6}&alarm/set\_alarm& V&\colorbox{RosyBrown1}{datetime}\\\hline
 45/2& Skal d bli \colorbox{LightBlue1}{sol} \colorbox{RosyBrown1}{i dag?}& weather/find& N&\makecell[l]{\colorbox{LightBlue1}{weather/attribute} \\\colorbox{RosyBrown1}{datetime}}\\\hline
 183/2& \makecell[l]{Æ vil gje \colorbox{DarkSeaGreen1}{boka} \colorbox{LightSteelBlue1}{3}\\ \colorbox{Burlywood1}{stjenre}.} & RateBook& N&\makecell[l]{\colorbox{DarkSeaGreen1}{object\_type}\\\hline\colorbox{LightSteelBlue1}{rating\_value}\\\colorbox{Burlywood1}{rating\_unit}}\\\bottomrule
    \end{tabular}
    \end{adjustbox}
    \caption{Random examples from the NoMusic development set.}
    \label{tab:examples}
\end{table}
}

\subsection{Initial Data: NoMusic Dataset}
\label{sec:datanomusic}
The shared task uses the NoMusic dataset \citep{maehlum-scherrer-2024-nomusic}, a ``multi-parallel resource for written Norwegian dialects, and the first evaluation dataset for slot and intent detection focusing on non-standard Norwegian varieties.''

To construct the development and test set (3300/5500 instances each), 11 Norwegian translators manually translated phrases from the corresponding English xSID sets \citep{van-der-goot-etal-2020-cross} into four different Norwegian dialects (North Norwegian, Trøndersk, West Norwegian and Bokmål). Shared task participants only had access to the dev set during the competition. See Table \ref{tab:examples} for examples from the dev set and Table \ref{tab:norsid-dev-test} for the distribution of labels.

For training data, a machine-translated version of the English xSID train set (43,605 instances) was provided.\footnote{More details of xSID are presented in Section \ref{sec:xsid}.} The instances have been translated into Bokmål and are annotated for both the intent detection and slot filling tasks. It preserves the original intent labels and the slots have been projected from one language to the other, although the shared task organizers report that the quality of both the translation and the annotation projection is relatively poor.

\begin{table}[ht]
\centering
\small
\begin{adjustbox}{max width=\linewidth}
\begin{tabular}{lrrr}
\toprule
Dialect          & Dev          &Test& Dist     \\
\midrule
West Norwegian (V)                & 1,500                    &2,500& 45.45\%            \\
 Trøndersk (T)                & 900                      & 1,500&27.27\% \\
North Norwegian (N)                & 600                      &1,000& 18.18\% \\
Bokmål (B)                & 300                      &500& 9.09\%  \\
\midrule
 Total& 3300& 5500&100\%\\
 \bottomrule
\end{tabular}%
\end{adjustbox}
\caption{Distribution of dialect tags in the NorSID development and test sets. Notice that the data distribution is highly skewed towards West Norwegian.}
\label{tab:norsid-dev-test}
\end{table}

\section{Intent Detection \& Slot Filling}
In this section, we will detail our participation in the intent and slot filling subtasks. We first explain the data (Section \ref{sec:xsid}) and the experimental design (Section \ref{sec:sid-experiments}), and finally a description and an analysis of our results (Section \ref{sec:sid-results}).

\subsection{Data}
\label{sec:xsid}

xSID \citep{van-der-goot-etal-2020-cross,2023-findings-vardial,Winkler2024} is a cross-lingual corpus for SLU.\footnote{As of version 0.6, the latest version to date, it is available in 17 languages.} The original English data was sampled by selecting random instances from the Snips \citep{coucke2018snipsvoiceplatformembedded} and Facebook \citep{schuster-etal-2019-cross-lingual-facebook} datasets. It features annotations for both intent detection, with one intent per instance; and slot filling, using the BIO format to tag each token. For the validation and test sets, the data was manually translated by native speakers of each language, maintaining the original intents, while the slots were manually re-annotated. The training data is available for most of the xSID languages through machine translation and projection of the slots.

For the Intent Detection task, there are a total of 18 intents. As per the slot filling task, there are 33 possible slots that can appear as the beginning (B) or inside (I) of a span and an O tag for the absence of entity. This results in a total of 67 possible tags.

Although the original paper leaves duplicated sentences to model the natural distribution found in the data, we deduplicate to avoid our models overfitting on the training data. We only carry out shallow deduplication, removing instances that contain the same text.

\subsection{Experiments}
\label{sec:sid-experiments}

\begin{figure}[ht]
    \centering
    \includegraphics[width=\linewidth]{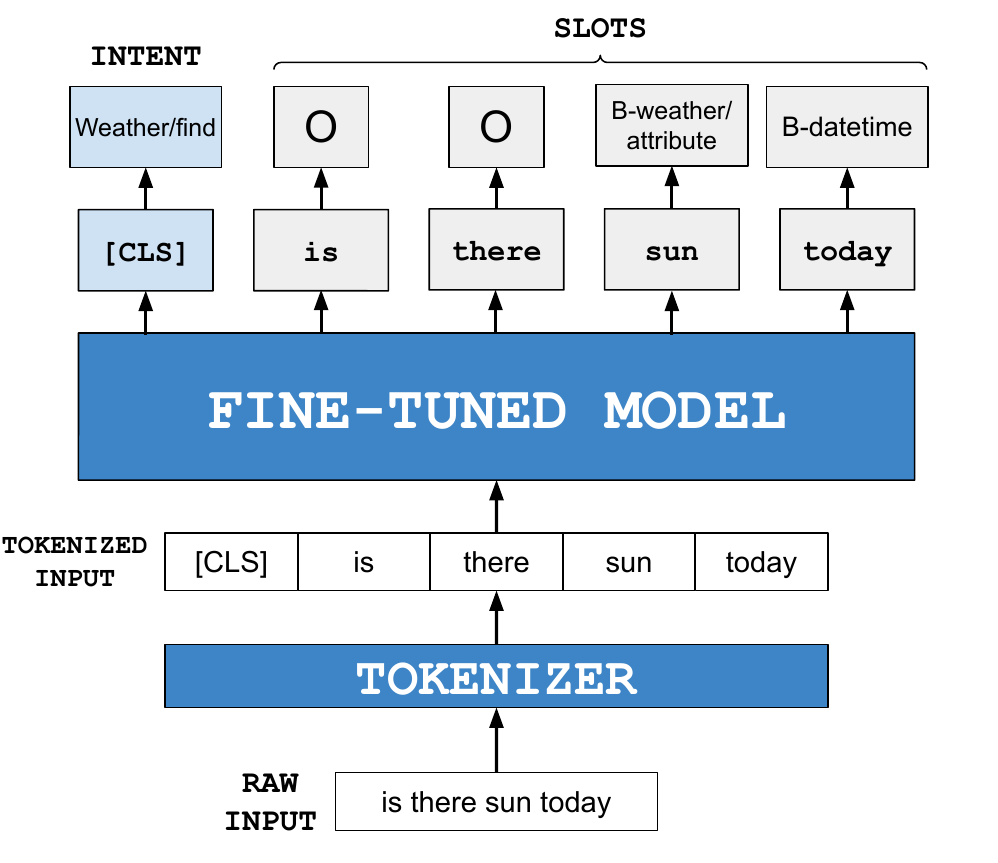}
    \caption{The idea behind the multitask model fine-tuned for both intent detection and slot filling tasks at the same time.}
    \label{fig:combined-model}
\end{figure}

Intent detection and slot filling are two highly related tasks. In fact, there are some slots that will only appear in sentences tagged with a certain intent and vice-versa. In this respect, a model could make use of the annotations of both tasks at the same time to obtain better predictions. Our experiments for the SID tasks build on that idea, using a multilingual multitask model jointly trained for intent detection and slot filling. As shown in Figure \ref{fig:combined-model}, our multitask models learn to classify the intents on top of the [CLS] token and the probabilities for each token on top of them.

Since intent detection and slot filling are classification tasks, we fine-tune the multilingual encoder model XLM-RoBERTa large \citep{xlm-roberta}. This allows us to take advantage of cross-lingual transfer by training on different combinations of languages from xSID.

The multitask loss is calculated as the weighted sum of the loss for intent and slot detection

\begin{equation} \label{eq:totalloss}
    \mathcal{L}_{total} =  \mathcal{L}_{slot} * \lambda + \mathcal{L}_{intent} * (1-\lambda)
\end{equation}

where $\mathcal{L}_{slot}$ is the cross-entropy loss function used in the slot-filling and $\mathcal{L}_{intent}$ is the cross-entropy loss function used in intent-detection. We set $\lambda$ to 0.7 based on the intuition that slot filling is more difficult.
The rest of the fine-tuning hyperparameters can be found in Appendix \ref{app:hyperparameters}.
During our experiments, all models have been evaluated using the Norwegian development set, which has been used to select the best combinations of languages and number of epochs.

\subsection{Results}
\label{sec:sid-results}

We have performed preliminary experiments on the development dataset to select the best combination of languages using three different random seeds. The results of these experiments can be seen in Table \ref{tab:slot_intent_dev_results}, where we report the \F and accuracy metrics for the slots and intents respectively \footnote{During the preliminary experiments on the development split, we have used a different scorer than the one provided by the Shared Task. Our scorer uses the output data of the model without post-processing, that allows us to calculate the scores while training.}. We also calculate the Lambda average metric, that is a weighted average, where we use the same $\lambda$ value as in the multitask loss function. 

\begin{table}[ht]
\centering
    \small
    \begin{adjustbox}{max width=\linewidth}
\begin{tabular}{lrrrrrrr}\toprule
Language &\F Slot &Accuracy Intent &Lambda \\\midrule
EN &\textbf{79.09} ${\scriptstyle\pm}$\scriptsize{0.77} &98.64 ${\scriptstyle\pm}$\scriptsize{0.23} &\textbf{84.96} ${\scriptstyle\pm}$\scriptsize{0.48} \\
DA &53.75 ${\scriptstyle\pm}$\scriptsize{0.35} &98.82 ${\scriptstyle\pm}$\scriptsize{0.56} &67.04 ${\scriptstyle\pm}$\scriptsize{0.05} \\
NB &53.49 ${\scriptstyle\pm}$\scriptsize{1.70} &98.87 ${\scriptstyle\pm}$\scriptsize{0.39} &67.10 ${\scriptstyle\pm}$\scriptsize{1.16} \\
EN+DA &57.03 ${\scriptstyle\pm}$\scriptsize{0.48} &98.60 ${\scriptstyle\pm}$\scriptsize{0.14} &69.50 ${\scriptstyle\pm}$\scriptsize{0.30} \\
EN+NB &55.43 ${\scriptstyle\pm}$\scriptsize{0.24} &\textbf{99.17} ${\scriptstyle\pm}$\scriptsize{0.13} &68.55 ${\scriptstyle\pm}$\scriptsize{0.20} \\
DA+NB &54.58 ${\scriptstyle\pm}$\scriptsize{0.32} &98.94 ${\scriptstyle\pm}$\scriptsize{0.21} &67.89 ${\scriptstyle\pm}$\scriptsize{0.18} \\
EN+DA+NB &58.33 ${\scriptstyle\pm}$\scriptsize{1.85} &98.73 ${\scriptstyle\pm}$\scriptsize{0.25} &70.45 ${\scriptstyle\pm}$\scriptsize{1.35} \\
ALL &59.83 ${\scriptstyle\pm}$\scriptsize{1.88} &98.67 ${\scriptstyle\pm}$\scriptsize{0.39} &71.48 ${\scriptstyle\pm}$\scriptsize{1.22} \\
ALL-NB &59.08 ${\scriptstyle\pm}$\scriptsize{0.83} &98.55 ${\scriptstyle\pm}$\scriptsize{0.41} &70.92 ${\scriptstyle\pm}$\scriptsize{0.66} \\
GER &58.01 ${\scriptstyle\pm}$\scriptsize{0.72} &98.80 ${\scriptstyle\pm}$\scriptsize{0.13} &70.25 ${\scriptstyle\pm}$\scriptsize{0.47} \\
GER-NB &61.96 ${\scriptstyle\pm}$\scriptsize{1.25} &98.25 ${\scriptstyle\pm}$\scriptsize{0.36} &72.85 ${\scriptstyle\pm}$\scriptsize{0.98} \\
LAT &58.51 ${\scriptstyle\pm}$\scriptsize{0.45} &98.80 ${\scriptstyle\pm}$\scriptsize{0.37} &70.60 ${\scriptstyle\pm}$\scriptsize{0.21} \\
LAT-NB &59.62 ${\scriptstyle\pm}$\scriptsize{1.34} &98.56 ${\scriptstyle\pm}$\scriptsize{0.42} &71.30 ${\scriptstyle\pm}$\scriptsize{1.04} \\
\bottomrule
\end{tabular}
\end{adjustbox}
\caption{\F score in the development set for each training language combination, labeling the tokenized sentences. ISO 639-1 Language Codes are used for individual languages, while ALL means the combination of all available training languages, GER means Germanic languages, and LAT means languages written in Latin script. We also sometimes remove Norwegian, e.g., GER-NB would be all Germanic languages except Norwegian Bokmål (full explanation in Appendix \ref{app:language_combination}).}
\label{tab:slot_intent_dev_results}
\end{table}

\begin{table}[ht]
\centering
\small
\begin{tabular}{lrr}
\toprule
&Single-task &Multitask \\
\midrule
Slot \F &78.98 ${\scriptstyle\pm}$\scriptsize{0.28} &\textbf{79.09}${\scriptstyle\pm}$\scriptsize{0.77}\\
Intent Accuracy &98.42${\scriptstyle\pm}$\scriptsize{0.31} &\textbf{98.64}${\scriptstyle\pm}$\scriptsize{0.23} \\
\bottomrule
\end{tabular}
\caption{Comparison between the single-task models and the multitask one.}\label{tab:slot_intent_comparison}

\end{table}

The results show that training only on the English training data produces the best results, with a Lambda average of 84.96\%, probably because machine-translated data can introduce noise to the model. 

Table \ref{tab:slot_intent_comparison} compares the multitask and single-task slot-filling and intent classification models, trained in English with the same hyperparameters. We see that not only is multitask training more efficient than single-task training, but it is also able to maintain a similar or slightly better performance (0.11\% and 0.22\% higher Slot \F and Intent accuracy respectively).

For the participation in the shared task, we submit three models: a) the model fine-tuned only on English data b) the model fine-tuned with a combination of English and Norwegian, which obtained the best accuracy for the intent task (99.17\%), and c) the model fine-tuned with the combination of all Germanic languages (that have an available training set) minus Norwegian, which obtained the second best results overall (72.85\% Lambda average).

The test results are shown in Table \ref{tab:slot_intent_test_results}. Consistent with the evaluation of our models with the development set, the best resulting model is the one trained only using English, with a Lambda average of 88.65\%.  

\begin{table}[ht]
\centering
 \small
\begin{adjustbox}{max width=\linewidth}
\begin{tabular}{lccc}\toprule
Model &Slot \F &Accuracy Intent &Lambda \\\midrule
EN &\textbf{85.37} &96.29 &\textbf{88.65} \\
GER-NB &66.64 &97.11 &75.78 \\
EN+NB &55.66 &\textbf{97.69} &68.27 \\
\bottomrule
\end{tabular}
\end{adjustbox}
\caption{Results (slot \F, accuracy intent, and lambda average) of the three submitted runs evaluated on the test set. Best results in bold.}\label{tab:slot_intent_test_results}
\end{table}

\subsubsection{Analysis of the Intent Detection Results}
Without any hyperparameter tuning, most models obtain near 100\% accuracy in the intent detection task. This is likely because the data is from a reduced domain, where instances contain clear word-level features that let the model infer the label.

\begin{figure}[ht]
    \centering
    \includegraphics[width=\linewidth]{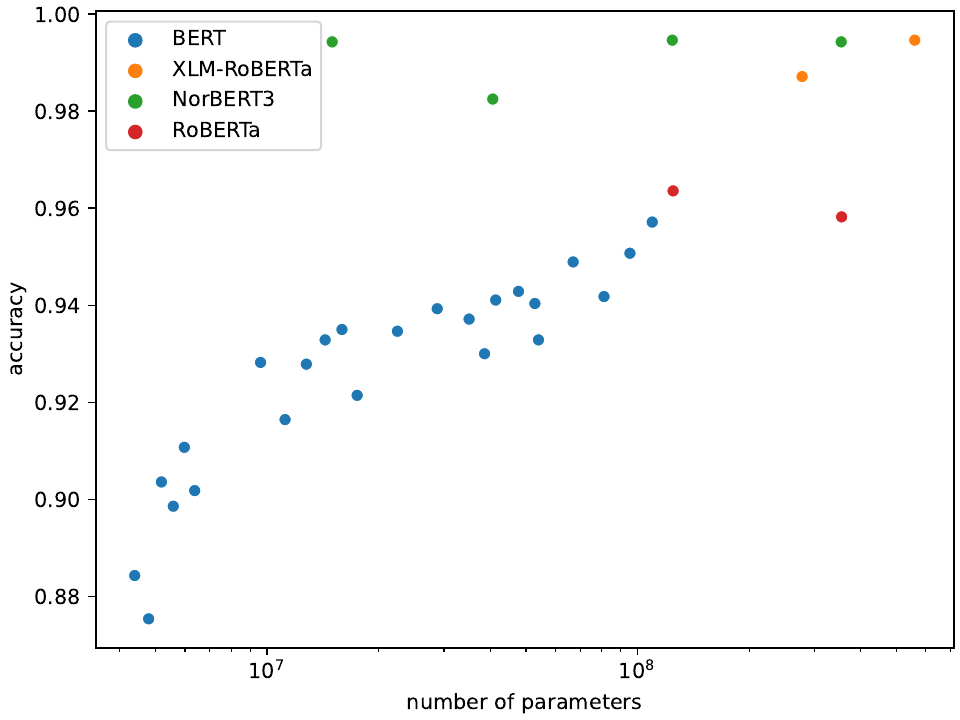}
    \caption{Accuracy of pretrained English models (BERT, RoBERTa), multilingual models (XLM-RoBERTa) and a Norwegian pretrained models (NorBERT3) trained for Intent Detection on the Norwegian train set and evaluated the development set.}
    \label{fig:ng_berts}
\end{figure}

To test this idea, we fine-tuned and compared the results of English only models, [BERT\footnote{\href{https://github.com/google-research/bert}{Google's 2020 BERT models} were fine-tuned.} \cite{devlin-etal-2019-bert, turc2019} and RoBERTa \cite{liu2019robertarobustlyoptimizedbert}], against multilingual and Norwegian models, [XLM-RoBERTa \cite{xlm-roberta} and NorBERT3 \cite{samuel-etal-2023-norbench}]. Figure \ref{fig:ng_berts} shows that no prior knowledge of Norwegian is required to obtain an accuracy of up to 96\%, which is aligned with our initial presumption that models are learning to classify the instances relying on specific word patterns rather than semantic understanding.
However, prior knowledge of Norwegian greatly reduces the number of parameters required to obtain top performance and allows the model to surpass the performance of English only models.

\section{Dialect Identification}

In this section, we will present the dialect identification task, starting with the data used in our experiments (Section \ref{sec:dialect_data}), followed by the experimental setting training only on the development set from the shared task (Section \ref{sec:dialect_experiments}), as well as the experimental settings when training on alternative sources of data (Section \ref{sec:dialect_additional}). Finally, we describe the results of using different data and settings (Section \ref{sec:dialect_results}).

\subsection{Data}
\label{sec:dialect_data}

The following section presents all the datasets we have used in our experiments, which consist of the NoMusic data (Table \ref{tab:norsid-dev-test}) , as well as some further dialectal data. This data comes from two main sources:  (i) tweets, which we collected from NorDial and the Nordic Tweet Stream (NTS); and (ii) transcriptions, which come from NB Samtale and the Nordic Dialect Corpus (NDC). 

\subsubsection{NoMusic}
\label{sec:nomusicsplit}
As introduced in Section \ref{sec:datanomusic}, NoMusic is the development data provided by the shared task. However, there is no additional training data that has been labeled for the dialect identification task in the SID tasks. 

Consequently, we split the development set into train, development and test sets (from now on, dev-train, dev-dev and dev-test). Each sentence in this dataset is paraphrased 11 times, once for each dialect annotator. Thus, in order to avoid data contamination, we split by the original ID of each instance, as many translated instances are similar or identical (Table \ref{tab:NoMusic-dev-splits}). The results presented in Section \ref{sec:dialect_experiments} correspond to the dev-test results. 

\begin{table}[ht]
\centering
\small
\begin{adjustbox}{max width=\linewidth}
\begin{tabular}{lrrr}
\toprule
Dialect          & Dev-Train          & Dev-Dev & Dev-Test     \\
\midrule
West Norwegian (V)                & 962                    &220& 225            \\
 Trøndersk (T)                & 580                      & 132&135 \\
North Norwegian (N)                & 386                      &89& 89 \\
Bokmål (B)                & 188                      &43& 45  \\
\midrule
 Total& 2116& 484&494\\
 \bottomrule
\end{tabular}
\end{adjustbox}
\caption{Distribution of splits in the development set.}
\label{tab:NoMusic-dev-splits}
\end{table}

\subsubsection{NorDial}
\label{sec:nordial}

NorDial \citep{barnes2021nordial} is a corpus of 1,073 Norwegian tweets annotated for four dialects: Bokmål, Nynorsk, Dialect, or Mixed. We merge this data together with the additional annotated data available in the Nordial GitHub,\footnote{\url{https://github.com/jerbarnes/nordial}} for a total of 6,670 tweets. \cref{tab:nordial-stats} shows the statistics for the merged data.

\begin{table}[ht]
\centering
\small
\begin{adjustbox}{max width=\linewidth}
\begin{tabular}{lrrrr}
\toprule
Split & Train & Dev & Test & Total \\
\midrule
Bokmål & 2798 & 115 & 98 & 3011 \\
Nynorsk & 964 & 38 & 43 & 1045 \\
Dialect & 2007 & 61 & 70 & 2138 \\
Mixed & 445 & 12 & 19 & 476 \\
\midrule
Total & 6214 & 226 & 230 & 6670 \\
\bottomrule
\end{tabular}
\end{adjustbox}
\caption{Distribution of Norwegian language variants across NorDial splits.}
\label{tab:nordial-stats}
\end{table}

\subsubsection{Nordic Tweet Stream (NTS)}
\label{sec:nts}

NTS\footnote{\url{https://nordictweetstream.fi/}} \citep{laitinen2018nordic} is a corpus of geolocated tweets and their associated metadata from the Nordic region between the years of 2013-2023. We downloaded 4.054.223 Norwegian tweets geolocated in a total of 426 Norwegian cities.

\subsubsection{NB Samtale}

NB Samtale\footnote{\url{https://huggingface.co/datasets/Sprakbanken/nb_samtale}} is a speech corpus collected by the Language Bank at the National Library of Norway. It contains orthographic and verbatim transcriptions from podcasts and recordings of live events at the National Library, a total of 24 hours of transcribed speech from 69 speakers, divided into train, development and test splits. \cref{tab:nb-samtale} shows the distribution of dialects in the data.

\begin{table}[ht]
\centering
\small
\begin{tabular}{ l r r r r}
\toprule
Dialect area& Train& Dev & Test&Total\\
\midrule
Eastern (E) & 4454&  557& 557& 5568\\
Northern (N) & 2072& 258& 261& 2591\\
Southwest (SW) & 1304&  164& 163& 1631\\
Western (W) & 1094&  137& 136& 1367\\
Central (T) & 624&  78& 78& 780\\
\midrule
Total& 9548& 1194& 1195& 11937\\
\bottomrule
\end{tabular}
\caption{Distribution of Norwegian language variants in NB Samtale.}
\label{tab:nb-samtale}
\end{table}

\subsubsection{Nordic Dialect Corpus (NDC)}
\label{sec:ndc}

NDC\footnote{\url{https://tekstlab.uio.no/scandiasyn/download.html}} \citep{johannessen2009nordic,johannessen2012nordic} includes orthographic and phonetic transcriptions of Nordic speaker recordings, with almost two million words from Norwegian dialects. It contains recordings from 111 different locations in Norway, collected between 2006-2010.

\subsection{Experiments With Development Data}
\label{sec:dialect_experiments}

In this section, we describe baselines using only the dialectal data in the development set, using the splits described in Section \ref{sec:nomusicsplit} (Table \ref{tab:NoMusic-dev-splits}). We explore lexical mapping SVM, fine-tuning encoders and decoders, as well as using few-shot decoders.

\subsubsection{Lexical Mapping SVM}
\label{sec:lex_map}

We first create a simple baseline by mapping common lexical items in Bokmål to their respective dialectal counterparts. The items we map are mainly pronouns and interrogatives, as well as a few common prepositions, verbal forms, and time expressions. For each dialect, there is often a one-to-many mapping from Bokmål, as can be seen in Table \ref{tab:lex_map}. 

\begin{table}[ht]
    \centering
    \small
    \begin{tabular}{lllll}
      \toprule
    B & V & T & N & EN\\
    \midrule
      jeg   & eg, ej & æ, e & æ, å & `I' \\
    hva & ka & ka & ka &  `what'\\
         \bottomrule
    \end{tabular}
    \caption{Example of lexical mappings for B, V, T, N. The English translation is added in the final column.}
    \label{tab:lex_map}
\end{table}

After compiling the lexical mappings, we create a silver dataset (\textbf{Lexmap}) starting from the Bokmål train data provided. Specifically, we create a new instance each for V, T, and N by mapping any lexical item in our mapping dictionary to its dialectal variant, leading to a training dataset four times the size of the original.

We train a linear support-vector machine on unigram features using the silver train set (\textbf{Lexmap SVM}). We also train the same model on the silver train plus the dev-train data (\textbf{Lexmap + dev-train SVM}).

\begin{table*}[ht]
\centering
\small
\begin{adjustbox}{max width=\linewidth}
\begin{tabular}{lcccccccc}
\toprule
 & \multicolumn{2}{c}{NDC-20 (ortho)} & \multicolumn{2}{c}{NDC-20 (phonetic)} & \multicolumn{2}{c}{NDC-40 (ortho)} & \multicolumn{2}{c}{NDC-40 (phonetic)} \\
\midrule
Dialect & \# & \% & \# & \% & \# & \% & \# & \%\\
\midrule
 West (V) & 31017 & 30.34 & 35636 & 30.84 & 21413 & 30.74 & 25803 & 31.28\\
 North (N)& 31387 & 30.70  &  34437 & 29.80 & 21487 & 30.85 & 24459 & 29.65\\
 Trøndersk (T) & 12076 & 11.81 &  13502 & 11.68 & 7989 & 11.47 & 9373 & 11.36\\
 Bokmål (B) & 27763 & 27.15 &  31990 & 27.68 & 18751 & 26.93 & 22868 & 27.72\\ 
 \midrule
 Total & 102243 & 100 &  115565 & 100 & 69640 & 100 & 82503 & 100\\
\bottomrule
\end{tabular}
\end{adjustbox}
\caption{Distribution of dialects in NDC, using a manual geolocation-based mapping of dialect labels, with a minimum token length of 20 and 40 per sentences}
\label{tab:dialectal-data}
\end{table*}

\begin{table*}[ht]
\centering
\small
\begin{adjustbox}{max width=\linewidth}
\begin{tabular}{lcccccccccc}
\toprule
 & \multicolumn{2}{c}{NDC-20 (ortho)} & \multicolumn{2}{c}{NDC-20 (phonetic)} & \multicolumn{2}{c}{NDC-40 (ortho)} & \multicolumn{2}{c}{NDC-40 (phonetic)} &  \multicolumn{2}{c}{NTS}\\
\midrule
Dialect & \# & \% & \# & \%  & \# & \% & \# & \% & \# & \%\\
\midrule
 West (V) & 6891 & 61.83 & 36418 & 37.24 & 5048 & 69.62 & 27845 & 38.72 & 49801 & 46.96\\
 North (N) & 52 & 0.47 & 218 & 0.22 & 33 & 0.46 & 45 & 0.06 & 16632 & 15.68\\
 Trøndersk (T) & 3877 & 34.79 & 60995 & 62.37 & 1976 & 27.25 & 43909 & 61.06 & 30007 & 28.30\\
 Bokmål (B) & 325 & 2.92 & 163 & 0.17 & 194 & 2.68 & 115 & 0.16 & 9609 & 9.06\\ 
 \midrule
 Total & 11145 & 100 & 97794  & 100 & 7251 & 100 & 71914 & 100 & 106049 & 100\\
\bottomrule
\end{tabular}
\end{adjustbox}
\caption{Distribution of dialects in NDC transcription and NTS tweet datasets, using automatic annotation of dialect labels and dropping instances to match the development distribution.}
\label{tab:transcription-data-automatic}
\end{table*}

\subsubsection{Encoder Fine-tuning}
\label{sec:encoders}
We fine-tune encoders on the \textbf{dev-train} set, as well as on the combination of dev-train with the lexical mapping silver (\textbf{Lexmap + dev-train)}. We choose the best encoder model specifically trained for Norwegian, NorBERT3-L \citep{samuel-etal-2023-norbench}, as well as the multilingual encoder model XLM-Roberta-large \cite{xlm-roberta}.\footnote{Hyperparameters used are listed in \cref{app:hyperparameters}.} As preliminary experiments showed training on the full development set with NorBERT3-L leads to the best performance, we also train the following variants: (i) training on the combined dev-train and dev-dev splits (\textbf{Dev-train-dev}); and (ii) training on the whole development set (\textbf{Dev-train-dev-test}). 

\subsubsection{Decoder Few-shot}
\label{sec:decoder_few_shot}

We perform few-shot prompting experiments, providing the model 4 example instances, one for each dialect label. The few-shot examples are sampled from the dev-dev split and we evaluate on the dev-test set. We experiment with a decoder model specifically trained for Norwegian, NorMistral-7b-warm,\footnote{\url{https://huggingface.co/norallm/normistral-7b-warm}} and a multilingual decoder model, Llama 3.1-8B \cite{dubey2024llama}, and use both base and instruct models, evaluating with LM evaluation Harness \cite{eval-harness}. The prompt used in these experiments is shown below:

\texttt{In which dialect is this text written? Choose between North Norwegian, Trøndersk, West Norwegian or Bokmål.
Text: \{text\} Dialect:}

\subsubsection{Decoder Fine-tuning}
\label{sec:decoder_finetuning}

Next, we fine-tune several decoders on the development set, similar to the experiments with decoders. We only experiment with NorMistral models, as they achieve higher results in few-shot evaluation. We perform finetuning in two ways: by adding a sequence classification (SC) head and training the models applying supervised fine-tuning (SFT) using the same English prompt as in the few-shot evaluation (\textbf{dev-train SFT}).

\subsection{Experiments With Other Data Sources}
\label{sec:dialect_additional}
As no labeled training dataset is available for dialect classification, we also explore whether it is possible to use other sources of data to learn to classify Norwegian dialects.

First, we apply the semi-automatic and automatic annotation methods (see subsections \ref{sec:semi-autom-annotation} \& \ref{sec:automatic_annotation}), and get statistics about the resulting dialectal distribution of tweets and transcriptions.

Next, we fine-tune NorBERT3-L on the semi-automatically and automatically labelled transcriptions and tweets to measure the impact of using automatically labeled data sources. During training, we use the dev-dev split as validation to avoid overfitting on these datasets and use the same hyperparameters (see \cref{app:hyperparameters}).

\subsubsection{Semi-automatic Annotation}
\label{sec:semi-autom-annotation}
We perform a semi-automatic dialect label annotation on the NDC dataset, by first eliminating special transcription characters, e.g., pause markers (\#) or (mm), as well as short sentences, which we assume have fewer dialectal traits.\footnote{We experiment with two different minimum sentence lengths: 20 and 40 tokens.} 

Finally, we semi-automatically map cities in NDC to their corresponding dialect label, according to their geographical location.\footnote{Eastern cities are mapped to Bokmål.} Table \ref{tab:dialectal-data} reports the number of instances and the dialect label distribution.

\subsubsection{Automatic Annotation}
\label{sec:automatic_annotation}

We automatically annotate silver training data using two classifiers: the best model trained on development data (see Section \ref{sec:results-dev}) and a model trained on NorDial data. Experiments on NorDial suggest NB-BERT-base is the strongest classifier, achieving 90\% weighted \F score, thus being chosen as our NorDial classifier. The objective of using two classifiers is to minimize model bias.

Therefore, having the results of our two classifiers, we discard examples classified as Nynorsk and Mixed by the NorDial classifier. For Bokmål, we select examples where the two classifiers match. For the dialectal tweets, we assign the class of the NorBERT3-L classifier if it is one of N, V or T.

\paragraph{NB Samtale}
We train a classifier on NB Samtale data with the available splits to measure to what extent there are dialectal features in the orthographic and verbatim transcriptions. We get a  weighted \F of 76.76\% with the verbatim transcriptions, so we can conclude that the models are able to learn the different features of the dataset. However, as training on this data leads to poor results on the dev set, we decide to explore other annotation methods. The poor results suggest that the dialectal features present in both datasets are different. Additionally, we trained a model using both NB Samtale train set and dev-train, but the results obtained (\F 81.59\%) are few points worse than the model trained only in dev-train (\F 82.44\%).

\paragraph{NTS} The predicted distribution of dialects in NTS tweets does not match with the Norsid classifier distribution. Nordial classifier classifies 96.70\% of instances as Bokmål and Norsid classifier 66.93\% as V. This makes sense because the distributions of their training data are different. After performing the automatic labeling, in order to obtain a distribution similar to the one we have in development, we have downsampled the automatically-labelled NTS instances until the distribution matches that of development (see \cref{tab:transcription-data-automatic}).

\paragraph{NDC}

We have additionally automatically annotated the NDC instances (see \cref{tab:transcription-data-automatic}). In most cases, there is a large difference between semi-automatic and automatic labeling. This could be due to the training data for our classifier differing from the instances in the NDC dataset, but we decided to follow the same annotation approach in order for the results to be comparable. Moreover, it is important to note that the automatic labeling distribution does not match the development set distribution; thus, our procedure has a bias toward annotating instances as V or T. 
The dialect identification results when using data annotated with this approach obtains better results than semi-automatic annotation and NB Samtale (see \cref{tab:dialect_results}), so we apply this classification method to the following dataset annotations.

\begin{table}[ht]
\centering
\small
\begin{adjustbox}{max width=\linewidth}
\begin{tabular}{ l  l  l l }
\toprule
Dataset & Model & Dev \F & Test \F \\
\midrule
\multirow{2}{*}{-} & Majority & 28.10 & 27.67 \\
& Random & 30.38 & 32.40 \\
\midrule
Lexmap & \multirow{2}{*}{SVM} & 53.91 & 56.11\\
Lexmap + dev-train & & 66.98 & 70.02 \\
\midrule
\multirow{2}{*}{Dev-train} & XLM-R-L & 61.85 & 63.76\\
& NorBERT3-L & \textbf{82.44} & 82.71\\
\midrule
Lexmap + dev-train & NorBERT3-L & 75.85 & 75.32 \\
\midrule
Dev-train-dev & \multirow{2}{*}{NorBERT3-L} & - & \textbf{84.17}\\
Dev-train-dev-test & & - & 83.34\\
\midrule
\multirow{4}{*}{Dev few-shot} & NorMistral-7b & 29.69 & 29.55 \\
& NorMistral-7b-it & 38.24 & 30.83\\
& Llama3.1-8B & 28.65 & 30.12 \\
& Llama3.1-8B-it & 28.64 & 28.88 \\
\midrule
\multirow{3}{*}{Dev-train} & NorMistral-7b (SC) & 78.69 & 74.91\\
& NorMistral-7b (SFT) & 76.79 & 76.88\\
& NorMistral-7b-it (SFT) & 76.43 & 74.16\\
\midrule
NTS* & NorBERT3-L & 64.60 & 64.22 \\
\midrule
NDC-20-orth* & \multirow{4}{*}{NorBERT3-L} & 33.65 & 34.10 \\
NDC-40-orth* & & 34.31 & 33.82 \\
NDC-20-phon* & & 51.23 & 52.09 \\
NDC-40-phon* & & 48.26 & 48.50 \\
\midrule
NDC-20-orth$\dagger$  & \multirow{4}{*}{NorBERT3-L} & 36.02 & 36.05 \\
NDC-40-orth$\dagger$ & & 32.08 & 35.39 \\
NDC-20-phon$\dagger$  & & 44.40 & 44.15 \\
NDC-40-phon$\dagger$  & & 44.97 & 43.78\\
\midrule
NB Samt & \multirow{2}{*}{NorBERT3-L} & 32.45 & 30.48\\
NB Samt + Dev-train & & 81.59 & 81.76\\
\bottomrule
\end{tabular}    
\end{adjustbox}
\caption{Weighted \F results of Dialect Identification subtask. * refers to the dataset annotated automatically and $\dagger$ to semi-automatically. \textit{it} refers to the instruct version of the models and \textit{L} the large version of the models.}
\label{tab:dialect_results}
\end{table}

\subsection{Results}
\label{sec:dialect_results}
The results were calculated using the official evaluation script of the shared task and the official metric, Weighted \F Score. All dev results in this section correspond to dev-test.%

\subsubsection{Training Only on Development Data}
\label{sec:results-dev}
The lexical mapping baseline performs better than majority or random, achieving 53.91 and 56.11 weighted \F on the dev-test and test sets, respectively. Further training on the dev-train set improves this to 66.98 and 70.02.

There is a large difference between the two encoder models (see \cref{tab:dialect_results}). Whereas XLM-Roberta does not reach the best lexical mapping baseline, NorBERT3-L surpasses the Lexmap + dev-train baseline by 15.46 points on the development set.  Additionally training with the Lexmap data, however, harms performance by 7 points. NorBERT3-L models trained in Dev-train-dev and Dev-train-dev-test obtain the highest results the test set.

In the few-shot scenario, the four models barely beat the majority class baseline (27.67) and perform worse than a random classifier (32.82). NorMistral Instruct (30.83) is slightly better than its base counterpart (29.55), but they are still far from the lexical mapping baseline, which obtains around 30 points more. Regarding Llama3.1 base and instruct models, their performance is almost identical to NorMistral models, but none of them surpass the performance of NorMistral Instruct in this few-shot evaluation. Fine-tuning NorMistral gives better results than the few-shot approach (76.88).

\subsubsection{Training on Other Sources of Data}

The results in \cref{tab:dialect_results} suggest that using tweets is better than transcriptions, in both semi-automatically and automatically labeled experiments: we obtain a weighted \F of 64.22 in our tweets model, while the transcription models perform between 30-52 points. However, the performance of the tweets model is still far from models trained on the development set (84.17).

When using transcriptions, the phonetic ones are preferable to orthographic ones, as more dialectal features are retained. Using longer sentences (>40 tokens) generally has little impact on performance, except for automatically labeled phonetic transcriptions.

The model trained on NB Samtale dataset achieves lower scores than models trained on NDC and NTS. 
This seems to be due to a low overlap in dialectal features between the NB Samtale and the shared task data.

\subsubsection{Dialect Analysis}

We have selected the best performing models from each strategy to analyze the performance in each dialect. The models we have chosen are, Dev-train-dev NorBERT3-L, Few-shot NorMistral-7b-warm-it, NTS NorBERT3-L, Semi-automatic labeled NDC-20-phon NorBERT3-L and Automatic labeled NDC-20-phon NorBERT3-L (see \cref{tab:dialect_analysis}).

For the best models trained on dev (NorBERT3-L and NorMistral-7b-warm (SFT)) the label imbalance affects performance, with models performing better on labels with more examples. We see this same pattern in the tweets dataset, as the dialect label distribution in the NTS dataset is similar to the one in the development set. For semi-automatic transcriptions, a higher performance is also observed on the majority classes, with the exception of Bokmål, probably due to annotation errors. In the automatic transcription datasets, the class imbalance is even larger, and this is reflected in even worse results for the minority classes. Finally, we see that the few-shot decoder model has a bias for T, as it assigns the other labels less often.

\begin{table}[ht]
    \centering
    \small
\begin{adjustbox}{max width=\linewidth}
    
    \begin{tabular}{cccccc}
    \toprule
        Dataset & Model & B & N & T & V\\
        \midrule
        Dev-train-dev & NorBERT3-L & 74.10 & 75.72 & 83.97 & 86.61\\
        Few-shot & NorMistral-7b-it & 06.56 & 00.88 & 42.12 & 12.87\\
        Dev-train &  NorMistral-7b (SFT) & 71.48 & 71.65 & 83.07 & 76.13 \\
        NTS & NorBERT3-L & 55.83 & 50.29 & 60.17 & 71.39\\
        NDC-20-phon$\dagger$ & NorBERT3-L & 14.17 & 39.73 & 19.95 & 58.75\\
        NDC-20-phon* & NorBERT3-L & 31.09 & 06.62 & 52.91 & 69.24\\
        \bottomrule
    \end{tabular}
    \end{adjustbox}
    \caption{Test \F per dialect with the best performing models in each category. \textit{it} refers to the instruct version of the models and \textit{L} the large version of the models.}
    \label{tab:dialect_analysis}
\end{table}

 \section{Conclusion and Future Work}
We have presented our submission for the NorSID Shared Task in the 2025 VarDial Workshop \cite{scherrer-etal-2025-vardial}. We have participated in the three proposed tasks -- Intent Detection, Slot Filling and Dialect Identification -- with 3 submissions for each of them.

For the Intent Detection \& Slot Filling tasks we designed a multitask model, improving efficiency with respect to having a model for each task. Additionally, as both tasks are highly related, this combination improves the performance of the model in both tasks to 97.69\% accuracy and 85.37\% \F, respectively, in the test set.

In Dialect Identification, we tested many different approaches by using the development data as training, as well as additional data from tweets and transcriptions. However, none of the settings we tried were able to surpass the performance of NorBERT3-L fine-tuned only on the development set, which achieved 84.17 \F on the test set.

The research presented in this paper has opened the way to many questions that need further investigation. We believe that the results could be improved using better encoder, e.g., DeBERTa \citep{he2021debertadecodingenhancedbertdisentangled}, and decoder, e.g., Llama 3.1 70B) models. The additional data we collected for dialect identification has not been successful due to the narrow domain of the tasks, but it is likely that for other tasks with a stronger domain shift this data could provide for more robust training.

\section*{Acknowledgements}

This work has been partially funded by:
\begin{itemize}
    \item Disargue (TED2021-130810B-C21) MCIN/AEI/10.13039/501100011033 and European Union NextGenerationEU/PRTR.
    \item DeepKnowledge (PID2021-127777OB-C21) MCIN/AEI/10.13039/501100011033 and by FEDER, EU.
    \item Ixa group A type research group (IT1570-22) Basque Government
    \item IKER-GAITU project 11:4711:23:410:23/0808 by Basque Government
    \item Julen Etxaniz holds a PhD grant from the Basque Government (PRE\_2024\_2\_0028).
    \item Maite Heredia is supported by the UPV/EHU PIF23/218 predoctoral grant.
    \item The EU Horizon Europe Framework under the grant 101135724 (LUMINOUS).
\end{itemize}

\bibliography{references}

\clearpage
\newpage
\appendix

\section{Hyperparameters}
\label{app:hyperparameters}
\subsection{Slot-intent Multitask Model}
The hyperparameters used in the slot-intent multitask model are the following:
\begin{itemize}
    \item \textit{Learning rate:} $2e^{-5}$
    \item \textit{Batch size:} $64$
    \item \textit{Number of epochs:} $10$
    \item \textit{Weight decay:} $0.01$
\end{itemize}
\subsection{Dialect Detection Model}

The hyperparameters used in dialect classification task are the following:

NorBERT3-L:
\begin{itemize}
    \item \textit{Learning rate:} $5e^{-5}$
    \item \textit{Batch size:} $16$
    \item \textit{Number of epochs:} $15$
    \item \textit{Weight decay:} $1e^{-4}$
\end{itemize}

XLM-RoBERTa-L:
\begin{itemize}
    \item \textit{Learning rate:} $1e^{-5}$
    \item \textit{Batch size:} $16$
    \item \textit{Number of epochs:} $15$
    \item \textit{Weight decay:} $1e^{-4}$
\end{itemize}

NorMistral:
\begin{itemize}
    \item \textit{Learning rate:} $5e^{-5}$
    \item \textit{Batch size:} $16$
    \item \textit{Number of epochs:} $5$
    \item \textit{Weight decay:} $1e^{-4}$
\end{itemize}

\section{Languages Combination}\label{app:language_combination}
The language combinations used in the slot-intent multitask model are the next ones: 
\begin{itemize}
    \item English (EN): Only the English language. This language is the only one that is not machine-translated in the xSID dataset
    \item Danish (DA): Only the Danish language. This language is the closest language to Norwegian in the xSID dataset.
    \item Norwegian (NB): Only the Norwegian training data provided. This data is poorly machine-translated, because of this, it was excluded from some combination of languages. 
    \item English and Danish (EN+DA): The combination of English and Danish languages.
    \item English and Norwegian (EN+NB): The combination of English and Norwegian languages.
    \item Danish and Norwegian (DA+NB): The combination of Danish and Norwegian languages.
    \item English, Danish, and Norwegian (EN+DA+NB): The combination of English, Danish, and Norwegian. 
    \item All languages (ALL): All languages on the xSID dataset (Arabic Danish German English Indonesian Italian Japanese Kazakh Dutch Serbian Turkish Chinese) and the Norwegian data provided. 
    \item All languages without Norwegian (ALL-NB): All languages on the xSID dataset (Arabic Danish German English Indonesian Italian Japanese Kazakh Dutch Serbian Turkish Chinese).
    \item Germanic languages (GER): Germanic languages on the xSID dataset (Danish German English Dutch) and the Norwegian data provided. 
    \item Germanic languages without Norwegian (GER-NB): Germanic languages on the xSID dataset (Danish German English Dutch) 
    \item Latin script languages (LAT): Languages that have latin script in the xSID dataset (Danish German English Indonesian Italian Dutch Serbian Turkish) and Norwegian.
    \item Latin script languages without Norwegian (LAT-NB): Languages that have latin script in the xSID dataset (Danish German English Indonesian Italian Dutch Serbian Turkish).
    
\end{itemize}

\end{document}